%% file: main.tex
\documentclass{bmvc2k}
\usepackage{import}
\usepackage{caption}
\usepackage{subcaption}
\usepackage{graphicx}
\usepackage{amsfonts}

\usepackage{hyperref}
\newcommand\tophline{\hline\noalign{\vspace{1mm}}}
\newcommand\middlehline{\noalign{\vspace{1mm}}\hline\noalign{\vspace{1mm}}}
\newcommand\bottomhline{\noalign{\vspace{1mm}}\hline}

\title{Variational Autoencoders for Feature Exploration and Malignancy Prediction of Lung Lesions}

\addauthor{Benjamin Keel}{ben.keel@gmail.com}{1} %https://eps.leeds.ac.uk/computing/pgr/9332/ben-keel
\addauthor{Aaron Quyn}{A.J.Quyn@leeds.ac.uk}{1,2} %https://baso.org.uk/about-us/national-committee/mr-aaron-quyn.aspx
\addauthor{David Jayne}{D.G.Jayne@leeds.ac.uk}{1,2} %https://medicinehealth.leeds.ac.uk/medicine/staff/475/professor-david-jayne
\addauthor{Samuel D. Relton}{S.D.Relton@leeds.ac.uk}{1} %https://medicinehealth.leeds.ac.uk/medicine/staff/706/dr-samuel-relton
\addinstitution{
 University of Leeds\\
 Leeds, UK
}
\addinstitution{
 Leeds Teaching Hospitals Trust\\
 Leeds, UK
}

\runninghead{Keel, Quyn, Jayne, Relton}{VAEs for Lung Lesion Malignancy Prediction}

\begin{document}

\maketitle

\begin{abstract}
Lung cancer is responsible for 21\% of cancer deaths in the UK and five-year survival rates are heavily influenced by the stage the cancer was identified at. Recent studies have demonstrated the capability of AI methods for accurate and early diagnosis of lung cancer from routine scans. However, this evidence has not translated into clinical practice with one barrier being a lack of interpretable models. This study investigates the application Variational Autoencoders (VAEs), a type of generative AI model, to lung cancer lesions. Proposed models were trained on lesions extracted from 3D CT scans in the LIDC-IDRI public dataset. Latent vector representations of 2D slices produced by the VAEs were explored through clustering to justify their quality and used in an MLP classifier model for lung cancer diagnosis, the best model achieved state-of-the-art metrics of AUC 0.98 and 93.1\% accuracy. Cluster analysis shows the VAE latent space separates the dataset of malignant and benign lesions based on meaningful feature components including tumour size, shape, patient and malignancy class. We also include a comparative analysis of the standard Gaussian VAE (GVAE) and the more recent Dirichlet VAE (DirVAE), which replaces the prior with a Dirichlet distribution to encourage a more explainable latent space with disentangled feature representation. Finally, we demonstrate the potential for latent space traversals corresponding to clinically meaningful feature changes. Our code is available at \url{https://github.com/benkeel/VAE_lung_lesion_BMVC}.
\end{abstract}

%-------------------------------------------------------------------------
\section{Introduction}
\label{sec:intro}
Lung cancer is the third most common cancer in the UK, 
accounting for 13\% of cases \cite{cancer_research_uk_2020} 
and the biggest cause of cancer death at 21\% \cite{cancer_research_uk_2020_mortality}. 
Early diagnosis of lung cancer is important for prognosis, 
with five-year survival rates for diagnosis in stages 1--3 at 32.6\% 
compared to 2.9\% at stage 4 \cite{five_year_survival}. 
Radiologists diagnose lung cancer from medical images including 
Computed Tomography (CT) scans by visually inspecting lesions in a 
time-consuming and subjective process \cite{lung_subjective}. 
A lesion is an area of tissue which has been damaged and is 
either a malignant tumour or a benign area of inflammation, 
abscess or ulcer \cite{national-cancer-institute}. 
CT scans are non-invasive and provide high detail images for 
medical diagnosis and treatment planning.
%Different parts of the body alter the X-ray beam in different ways, denser structures such as bone absorb more of the beam and so provide a contrast to less dense areas such as tissues that absorb less, providing an image.
The main contribution of this research is to:
\begin{itemize}
    \item Build state-of-the-art prediction models for lung cancer lesions using VAEs.
    \item Investigate the effectiveness of Dirichlet VAEs for lung lesions, to the best of our knowledge this is the first application in the cancer imaging domain.
\end{itemize}
%A first application of the DirVAE in the cancer imaging domain. 
Several research papers have investigated the application of AI methods to lung cancer, utilising their ability for complex pattern recognition \cite{cancers14061370, systematic_review}. 
The Variational Autoencoder (VAE) is an encoder-decoder architecture 
that maps input data to an $n$-dimensional latent space \cite{vae_original}. 
Smoothness constraints on the latent space, typically enforced using a Gaussian distribution, 
promote clustering between similar images. 
Assuming this space captures sufficient information, 
these latent vectors can be used for classification purposes. 
Exploration of the space via latent arithmetic and clustering can 
lead to new insights about a dataset \cite{higgins2017betavae, Klys2018, Shon2022}. 
%For instance, a latent space learned from images of tumours may link groups together by shape or texture features different from clinical diagnostic criteria. 
% Sam: Not sure we use the next sentence later so have provisionally removed for now.
%Finally, sampling from the latent space allows the generation of synthetic data through the decoder, either from randomly generated latent vectors or by using an informative direction in the latent space to change the latent vector in a way that manipulates features in a given image. Both may be useful for further classification tasks.

\begin{figure}
    \centering
    \includegraphics[width=\textwidth]{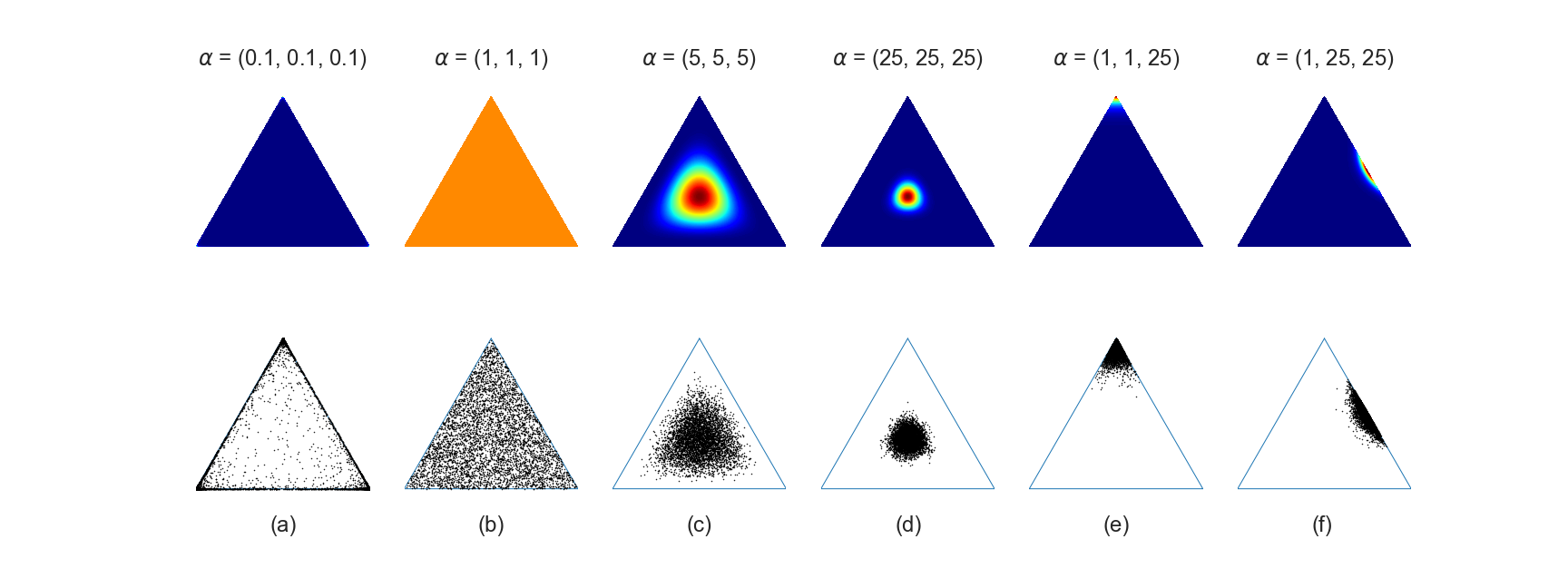}
    \caption{Examples of Dirichlet distribution simplex with 5000 samples given $\alpha$ parameters.}
    \label{fig:dirichlet_examples}
    %\vspace{-2mm}
\end{figure}

This paper also explores the use of a Dirichlet distribution in place of the Gaussian.
The $K$-dimensional Dirichlet distribution is a multivariate generalisation of the beta distribution with 
$K$ strictly positive parameters, 
$\left\{\alpha_i \in \mathbb{R}^{+} \right\}^{i=K}_{i=1}$. 
These $\alpha$ parameters influence the sparsity and density of the probability simplex, 
the impact of different values is shown in Figure \ref{fig:dirichlet_examples}. The sum of the $\alpha$ values is known as the concentration parameter, 
which controls the dispersion. When all $\alpha$ equal $1$ it is a uniform distribution (Figure \ref{fig:dirichlet_examples} (b)) and a lower/higher sum causes sparsity/density (Figure \ref{fig:dirichlet_examples} (a), (c), (d)). 
A relatively high $\alpha_i$ will encourage more probability to be 
concentrated in the corresponding area of the simplex (Figure \ref{fig:dirichlet_examples} (e), (f)). Choosing target $\alpha$ values in DirVAE influences the distribution of the VAE latent space.
 
%(b)), a sum close to $0$ will yield a higher density at the vertices meaning a more sparse representation, with most probabilities near zero (Figure \ref{fig:dirichlet_examples} (a)). Whereas, a very large sum ($>>K$) will yield a probability simplex which is more densely concentrated in the centre (Figure \ref{fig:dirichlet_examples} (c), (d)). 

% TODO: Come back and make this more succinct later
In summary, VAE models will be trained on 2D slices of CT scans cropped to lung lesions. The latent vector representations are used in Multilayered Perceptron (MLP) classification models for the task of lung cancer diagnosis. The latent vectors will be evaluated to justify their quality as feature vectors by showing that tumours with similar characteristics are grouped together in the latent space and to demonstrate the ability to predictably change features. This enhances the explainability of the method as it is more intuitive and interpretable for a non-technical audience. Additionally, comparisons between the Gaussian and Dirichlet latent space will show that the DirVAE has better disentanglement of features. To inform this research, we conducted a review of the published literature on AI for lung lesion diagnosis, applications of VAEs in the cancer domain and applications of DirVAEs.
%The code for this study is provided at \url{https://github.com/benkeel/VAE_lung_lesion}.

\section{Related Works} \label{lit_review}
% Sam: The point of this section is to find what's been done already and then state
% how our contributions build upon this - the latter part isn't in there yet.
\citet{systematic_review} conducted a systematic review of 
Artificial Intelligence (AI) for lung lesion diagnosis from 
medical images in the years 2017-2021 and found an accuracy range of 88\% to 99.2\% 
and AUC range of 0.7 to 0.967. 
Over half of the studies use 2D Convolutional Neural Network (CNN) architectures 
for feature extraction and a separate classifier, 
with transfer learning (TL) commonly applied. 
For instance, 
\citet{9076423} used TL with ResNet 50 \cite{DBLP:journals/corr/HeZRS15} 
and a shallow CNN, achieving 97.6\% accuracy. 
% Sam: Why is this interpretable?
Additionally, some studies have fused clinically known features with CNN derived features, for instance, \citet{xie} obtained AUC of 0.967 and an accuracy of 89.5\%. 
\citet{systematic_review} did not include any papers applying VAE to lung cancer detection, however, there are some existing studies in this domain \cite{silva, Astaraki2021, Gao2023}. In the most similar study with the best diagnostic performance using VAEs, \citet{silva} applied a VAE to lesions extracted 
from the LIDC-IDRI dataset 
and used retraining of the encoder with a Multi-Layered Perceptron (MLP) classifier, achieving AUC of 0.936. Additionally, several papers have applied VAEs to lung cancer for other tasks including segmentation, survival analysis and tumour growth prediction \cite{Pastor-Serrano2021, Li2022lung, Vo2020, Xiao2020, Jiang2021, Wang2019lung}.
This paper builds upon the work of \citet{silva} by improving both the diagnostic performance and the interpretability of the method.

%While many studies have applied AI methods to lung cancer diagnosis, 
%most lack appropriate validation and explainability which limit 
%their potential for clinical use. 
% Sam: This isn't a convincing argument of explanability (to me).

Regarding the application of generative models to the cancer domain,
several papers have explored the value of VAEs for latent space exploration \cite{Wang2019lung, latentspace_vaes, liver-latent}. For instance,
\citet{Wang2019lung} used VAEs to learn latent representations of the DNA to classify lung cancer subtypes.
%\citet{latentspace_vaes} trained a VAE and generative adversarial network (GAN) on CT scans from the LIDC-IDRI dataset and demonstrated they could be used to learn interpretable directions in the latent space for manipulating features. 
\citet{liver-latent} used an approach based on autoencoders and GANs for
generating synthetic abdominal CT scans and demonstrated adding and removing liver lesions. %add one more example here if there is space 

Several previous studies have proposed VAEs which replace the prior distribution with a Dirichlet. 
However, to our knowledge, our work is the first to apply this idea within a cancer setting.
%to the best of our knowledge none were within the cancer imaging domain. 
The DirVAE was originally proposed by \citet{Srivastava2017} and was subsequently 
utilised in similar studies on topic modelling by \citet{Xiao2018} and \citet{Burkhardt2019}. 
Later studies applied the model to image classification and demonstrated that 
DirVAE latent vectors were very capable in clustering images from the same category 
and separating them from others \cite{Joo2020,Gawlikowski2022}. 
\citet{Li2020} proposed an approach which combined graph neural networks and the 
DirVAE for abstract graph clustering. 
In the medical domain, 
\citet{Kshirsagar2022} used the approach to disentangle DNA sequences into different cell types. 
Most recently, \citet{Harkness2023} used the DirVAE for chest X-ray classification. %Notably, this work used the same reparametrisation trick as this paper.

%Several recent research papers have investigated the use of the Dirichlet as a prior distribution in a VAE, it can have the effect of encouraging sparsity and disentanglement in the latent space. However, it may also lead to a weaker generalisation of the data and lower quality reconstructions by the decoder when compared to a GVAE (see Results). As demonstrated in Figure \ref{fig:dirichlet_examples} (a), (e), (f), the distribution can promote sparsity where sampled values are closer to the vertices of the probability simplex. This promotes disentanglement of the latent space as the re-sampling technique will be more likely to produce inputs to the decoder with values concentrated in a single or small group of dimensions. This encourages the model to learn to produce a latent space which encodes features separately across the dimensions. In a medical context, this property may be useful for explaining the model to clinicians. For instance, demonstrating that changing a single number and re-sampling the tumour CT scan can lead to a larger tumour diameter or a more irregular border. 

% Sam: Possibly too much detail - have left in for now
Using the Dirichlet distribution in a VAE requires a reparameterisation trick which can produce a differentiable sample from the theoretical distribution. Various techniques have been used before which include the Laplace approximation \cite{Li2020}, approximation of the inverse CDF \cite{Joo2020}, rejection sampling variational inference \cite{Burkhardt2019} and implicit reparamterisation gradients \cite{Figurnov2018}. Instead, sampling from the Dirichlet distribution is done using the pathwise gradient method introduced in \cite{Jankowiak2018} and subsequently implemented in PyTorch.

%\cite{SUN2017530} with AUC 89.9\%
%\cite{svm} with accuracy 91.9\%
%\cite{PAUL2020103882} with AUC 0.96 and accuracy 90.29\%

%\cite{wang_et_al} 90.38\% accuracy and AUC 0.958  % not LIDC-IDRI

%Shen et al. proposed a method which used a CNN for feature extraction along with a random forest classifier of malignancy vs benign with accuracy of 86.8\% \cite{shen}. 

%A similar method was proposed in Lu et al., using CNNs for feature extraction but with a support vector machine (SVM) as the classifier, and achieved 91.9\% accuracy  \cite{svm}.

\section{Methods}
\subsection{Dataset and Pre-Processing} 
The LIDC-IDRI public dataset contains 1,010 CT scans,
consisting of 20,801 2D image slices which range from 0.6 to 5.0 mm thick 
with expert annotations \cite{lidc-paper, lidc-data}. 
The dataset was then limited to 875 patients with a lesion present totalling 13,916 slices. 
\citet{silva} reported that the LIDC-IDRI contains 2,669 lesions larger than 3 mm. 
The lesions are categorised as malignant, ambiguous or benign in 
5,249, 5,393 and 3,274 slices respectively, 
corresponding to 394, 580 and 454 patients. 
Note that some patients exhibit all three types. 
%The lesion group split is shown in Table \ref{table1},  
%\begin{table}[h!]
%\caption{Meta Data}
%\centering
%\begin{tabular}{|c|c|c|c|}\hline
%\bf{Status} & \bf{Number of Patients} & \bf{Number of Slices} & \bf{Proportion of %Dataset} \\ \hline
%Cancerous (Tumour) & 394 & 5249 & 37.72\% \\ \hline
%Ambiguous & 580 & 5393 & 38.75\% \\ \hline
%Non-Cancerous & 454 & 3274 & 23.53\%\\ \hline
%\end{tabular}
%\label{table1}
%\end{table}
%\newline \noindent
These labels were assigned based on a score of 1-5 agreed by four experienced thoracic radiologists: 
lesions with a score of 1 or 2 are benign, 3 is ambiguous, 
and 4 or 5 are malignant. 
All slices have segmentation masks that indicate where the lesion is located. 
Lesions measuring less than 3 mm in diameter and additionally 
any with less than 8 pixels were removed as they correspond to much smaller lesions 
which are not clinically relevant \cite{lesions4mm, lesions_rel}. 

Image slices are 512x512 pixels covering the cross-section of the body, from this a region of interest (ROI) of size 64x64 containing the segmentation masks was selected. Subsequently, 24 slices were excluded as they did not fit in the ROI and a further 64 slices as the bounding box went over the edge of the image, 
leaving a total of 13,852 in the final dataset. 
%The convex hull of each segmentation mask and subsequently fit the smallest bounding box around it. The largest of the minimum bounding boxes was 72x72, however since only 24 bounding boxes were larger than 64x64, $2^6$ was chosen for computational efficiency. The centre of the ROI for each slice was calculated using the formula for $x$ and $y$ on the accompanying mask, $\frac{1}{2}(\text{max}(x)-\text{min}(x))$.

%While it was possible to move the centre of the ROI, it would have led to systematically different samples. 
Pixels in the scan are dimensionless Hounsfield units (HUs) in the range\linebreak $[-3000,3000] \in \mathbb{R}$. 
HUs measure the intensity of an X-ray beam, which is altered based on the density of a structure. 
In this context, HU values below -1000 correspond to air, above 400 are bone, and in between are tissues. 
Since this work is concerned with lesions which are based in the tissues, upper and lower limits are set for the HU and values are scaled to the range $(0,1)$ as in \citet{silva}. 
This scaling will help to homogenise structures of bone and air to reduce variation.

\subsection{Model Description and Training}

\subsubsection{Initial VAE Training}
\begin{figure} %[h!] 
\centering % change to 4.5
\includegraphics[width=\textwidth]{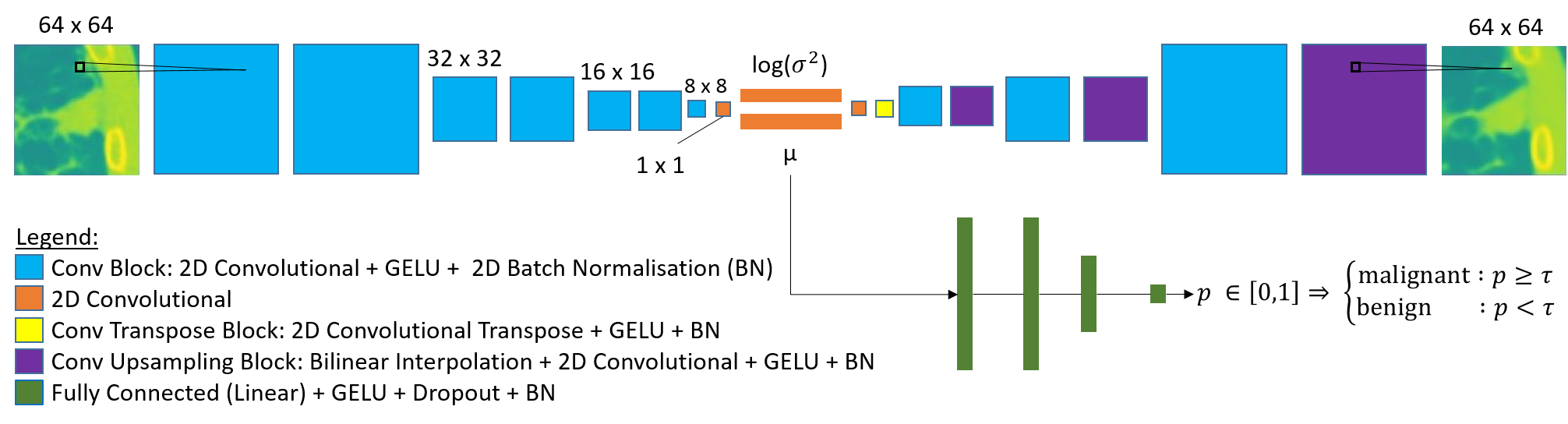}
\vspace{0.1cm}
\caption{Proposed Model Architecture: VAE with MLP classifier. The feature map size is in the format `$n$ x $n$' and these sizes follow left to right. The decoder architecture is symmetric to the encoder. Note that the final layers for the VAE and MLP have Sigmoid activation instead of GELU and do not have Batch Normalisation (BN) or Dropout.}
\label{model architecture}
%\vspace{-2mm}
\end{figure}

The VAE architecture proposed in this paper is visualised in Figure \ref{model architecture}. The architecture is loosely adapted from \cite{hansen} with additional hyperparameter training and different activation functions. The encoder component uses blocks of 2D Convolutional (Conv) layers with a Gaussian Error Linear Unit (GELU) activation function \cite{GELU} and 2D Batch Normalisation \cite{batch-norm}. For the Gaussian VAE, the output of the encoder is used in two separate 2D Conv layers for mean ($\mu$) and log variance $\left( \text{log}(\sigma^2) \right)$, whereas in the DirVAE a single 2D linear layer is used for the alpha ($\alpha$) parameters. These layers form a latent space of lesion feature representations for the respective models. The decoder takes a parameterised version of the latent vectors, sampled from an $n$-dimensional Gaussian or Dirichlet distribution. The decoder is a symmetric architecture which applies upsampling to the feature maps to reconstruct the images. Firstly, with a 2D Convolutional Transpose layer and secondly, using a combination of bilinear interpolation with 2D Conv layers. This second approach is less computationally expensive and helps avoid artifacts \cite{odena2016deconvolution}. The decoder produces a tensor of the same shape as the input containing the reconstructed images which are then evaluated against the original images in the loss function. 

%In the decoder, the feature map number and size are symmetric to the encoder.

%\begin{equation}
%    \mu + \alpha \cdot e^{\text{log}(\sigma^2)} = \mu + \alpha \cdot \sigma^2
%\end{equation}
%where $\alpha \sim N(0,1)$, so a maximum of $\sigma^2$ is added to each $\mu$. 

%This reparametrisation makes the latent space continuous, as each mean value has an n-d ellipse of possible points with radius equal to the variance. 
 
The loss function is a weighted combination of three terms: the L1 Loss, the Kullback-Leibler Divergence (KLD) \cite{kld} and the Structural Similarity Index Measure (SSIM) \cite{ssim} or the Multi-Scale SSIM (MS-SSIM) \cite{ms-ssim} for each image $i$ as follows,
%\vspace{-1mm}
\begin{equation} \label{loss function}
    \frac{1}{\text{batch\_size} \cdot \text{base}} \displaystyle\sum_{i=1}^n  \lambda \cdot \psi \cdot \text{L1 Loss}_i + (1-\lambda) \cdot \gamma \cdot \text{SSIM}_i + a \cdot \beta_{\text{norm}} \cdot \text{KLD}_i.
    %\vspace{-0.5mm}
\end{equation}
The scale factor ${(\text{batch\_size} \cdot \text{base})}^{-1}$ is applied so that the values are consistent across different hyperparameters; `base' is a scalar parameter controlling the 
number of feature maps in the VAE model.
The first two components, L1 Loss and either SSIM or MS-SSIM, measure image reconstruction quality and the KLD is the standard measure of latent space smoothness \cite{vae_original}. 
The reconstruction metrics are balanced using the hyperparameter constant 
$\lambda\in[0,1]$. 
Two other hyperparameters are used to weight theses components, 
$\psi \in \{1,2,3\}$ and $\gamma \in \{0,1, \text{batch\_size}\}$ 
which is used to either exclude or 
include the mean or the sum of the SSIM.
Finally, the KLD is scaled by the hyperparameter $\beta_{\text{norm}} = \beta \cdot \frac{\text{latent\_size}}{\text{image\_size}}$, as discussed in \citet{higgins2017betavae}, this formulation with $\beta>1$ leads to better disentanglement of the latent space, here $\beta$ values are in the range $[1,50]$. An annealing function $a$ was also included which linearly decreases the KLD by a maximum of 1 across the training epochs. 
%exponentially increases the weight from 1 to 3 across the training. 
The loss function was altered based on the above hyperparameters to find a combination which balanced the adversarial objectives of image quality and latent space smoothness. 

In total, 
the VAE models have 12 trainable hyperparameters which were 
explored using a random search strategy, 
including upper and lower bound for the HU, 
number of feature maps in VAE layers (base), 
size of the latent vector, 
the 4 parameters in the loss function in equation \ref{loss function}, 
whether to use the SSIM or MS-SSIM, whether or not annealing was applied to the 
KLD, the learning rate and batch size.\footnote{Code for this paper including hyperparameters used during the random search are available from the GitHub page: \url{https://github.com/benkeel/VAE_lung_lesion_BMVC} } 
The DirVAE had an additional hyperparameter for the target alpha parameters which the KLD compares against tries to move towards. The values are in the range $\alpha_i \in [0.5, 0.99]$, ranging from a sparse and disentangled distribution to almost a uniform distribution at higher values (c.f.~Figure \ref{fig:dirichlet_examples} (a) and (b)). 

The dataset of 875 patients was randomly split 70/30 into train and test sets with approximately 613 and 262 patients. The VAE reconstructions were evaluated qualitatively, and quantitatively with the average SSIM, Mean Squared Error (MSE) and Mean Absolute Error (MAE) which are conventionally used in the literature. 

\subsubsection{Fine-Tuning and Classification}
After initial training, the loss function \eqref{loss function} is updated 
to add a new term `$\text{BCE}_i$' which is the binary cross entropy loss
\cite{bridle1991probabilistic} of the MLP malignancy classifier \cite{mlp} shown in the model architecture (Figure \ref{model architecture}). The aim is to enable the VAE to be simultaneously useful for 
reconstruction and classification.

%and the Multi-Layer Perceptron (MLP)  shown in our model architecture (Figure~\ref{model architecture}) has not yet been utilised.

We employ a greedy optimisation strategy similar to 
Expectation-Maximimisation (EM) optimisation as described in the following pseudocode.
\begin{enumerate}
   \item Train the VAE model using loss function \eqref{loss function} and 
   extract the latent vectors.
   \item Using these latent vectors, find optimal hyperparameters for the MLP classifier using BCE loss.
   \item Repeat steps 1 and 2 until convergence, adding the BCE loss of the current optimal MLP to the loss function.
\end{enumerate}

The MLP hidden layers include GELU activation, dropout and batch normalisation,
with a sigmoid activation on the output layer to return probabilities,
with a parameter $\tau$ controlling the threshold beyond 
which a example is predicted as positive.
The key hyperparameters which were trained using a random search strategy 
include $\tau$ with a value in the range $[0.4,0.6]$, learning rate, 
batch size, number of nodes in each layer, 
whether there are 4 or 5 layers and a dropout probability.  
The 13,852 slices were split into 5 sets with 
train, validation and test sets in ratio $3:1:1$ for 
5-fold cross-validation; evaluation metrics are reported as the mean of these runs with standard deviations given for AUC and accuracy.
Classification performance will be evaluated using the AUC primarily, 
though we also report the accuracy, precision, recall, specificity, and F1-score.
% Note that we have labelled the lesions with an ambiguous label as benign but acknowledge that in practice these cases should be flagged to be followed up on. 
%Also note that the learning rate is reduced by a factor of 4 for this fine-tuning stage to help maintain the reconstruction quality and latent space smoothness. 
% Sam: Think it's OK to just have this in the Github.
The VAE and MLP models were built in Python 3.9 using PyTorch 1.12 and trained using the Adam optimiser \cite{adam}.

%\begin{equation}
%   \displaystyle\sum_{i=1}^n |x_{\text{true}} - x_{\text{predicted}}|
%\end{equation}
%firstly the l1 loss which is the sum of the absolute errors of the pixels in the reconstructed image compared to the original image
%\begin{equation}
%   \displaystyle\sum_{i=1}^n |x_{\text{true}} - x_{\text{predicted}}|
%\end{equation}

\subsubsection{Clustering and Latent Space Exploration} \label{section_clustering_explanation}
Two clustering methods, K-Means \cite{kmeans} and CLASSIX \cite{CG22b}, 
were used to partition the latent vectors into distinct groups. 
%To strike a good balance between the number of clusters and their density, 
An optimal range of values for parameter $k$, the number of clusters in K-Means, was investigated with an elbow graph of the 
sum of squared distances within each cluster to find a good balance 
in the number of clusters and their density. The density parameter in CLASSIX is chosen using a grid search to maximise separation by malignancy class. K-Means is non-deterministic and so results are averaged over 50 runs.
%which plots the number of clusters against the sum of squared distances for each point to the nearest cluster centre (centroid). The elbow of the graph, where the gradient shifts significantly, is considered the optimal k. The clusters will be investigated by qualitatively describing a random sample from the most benign and most malignant clusters for each set of latent vectors and by generate statistics which show the proportion of slices from the same patient in a cluster and the proportions of malignant or benign lesions.

% SAM: Would add something about latent space exploration here
Directions in the latent space corresponding to feature changes were found by collecting two groups of latent vectors, with and without a desired feature and taking the average direction vector between the groups. Latent traversal figures were produced by applying multiples of the direction vector to a new image and plotting the decoded images.
% e.g. big and small tumour

\section{Results}
\subsection{VAE Lung Lesion Reconstructions}
\begin{figure}[ht]
\centering
\begin{minipage}[b]{0.4\linewidth}
\centering
\includegraphics[width=\textwidth]{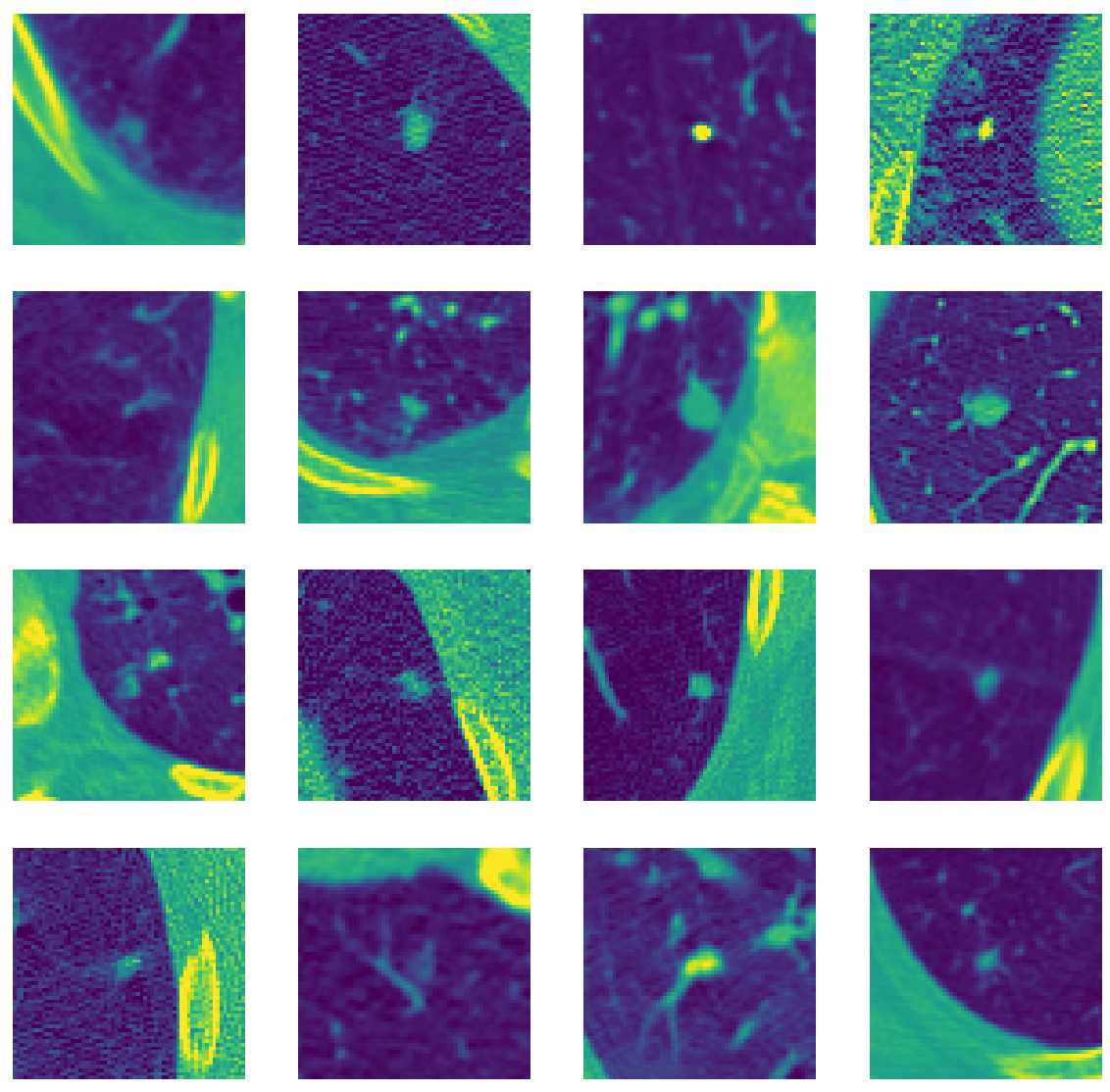}
%\subcaption{(a)}
\end{minipage}
\hspace{0.5cm}
\begin{minipage}[b]{0.4\linewidth}
\centering
\includegraphics[width=\textwidth]{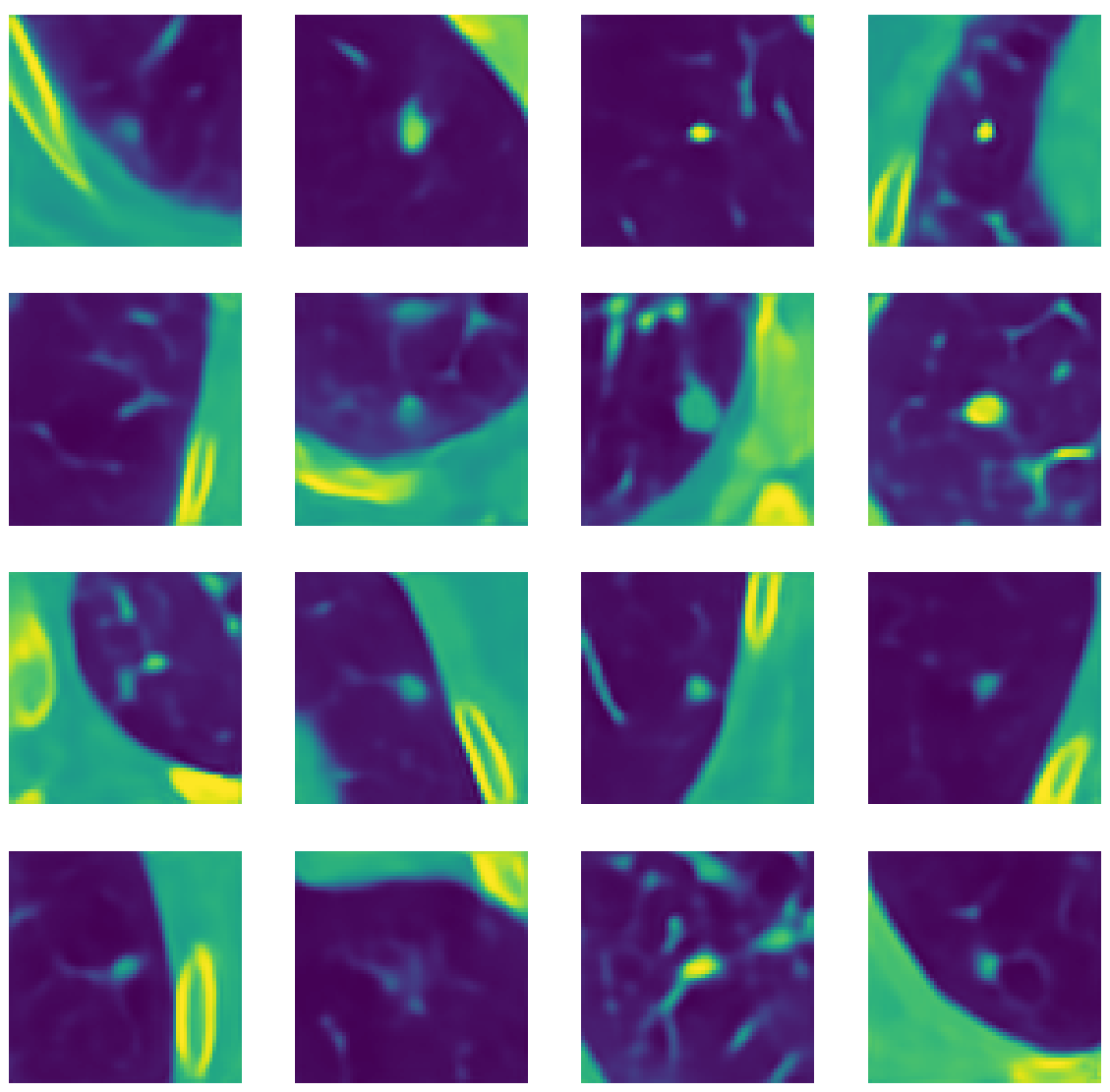}
%\subcaption{(b)}
\end{minipage}
\vspace{0.3cm}
\caption{Demonstration of VAE reconstruction quality with original images (a) and the corresponding VAE reconstructions (b).}
\label{vaerecon}
%\vspace{-2mm}
\end{figure}

Here a random sample of 16 images and the reconstructions by the GVAE are qualitatively reviewed in Figure \ref{vaerecon}. Firstly, observe that the overall macrostructure is captured well and so are most of the microstructures, however, some heterogeneity is lost. The most obvious missing information is that some of the lung parenchyma which could be alveoli are not fully captured in the reconstructed versions. Clinical collaborators specialising in oncology, AQ and DJ, 
confirmed the reconstructions captured the important clinical features 
considered in diagnosis. 
Based on a hyperparameter search of around 120 GVAE and 40 DirVAE candidates, %40 candidates for each, 
overall the DirVAE had a poorer image reconstruction. %but both were capable for this task and could produce realistic synthetic images. 
The best GVAE achieved SSIM of 0.89, MSE of 0.0032 and MAE of 0.027, 
whereas the best DirVAE achieved SSIM 0.65, MSE of 0.017 and MAE of 0.055.

% Loss of texture is in fatty tissue (is noise, had no relevance to diagnosis)

% don't care about random dots but do care about lines which connect tumour
% - re-watch recording for this
% difficult to say whether is is normal parenchyma or not - would take an expert opinion 
% - add this to discussion

%compare MSE of these reconstructions so we can compare to \cite{silva}

% frechet?

%-	may want to blank out normal lung but not losing speculation related to tumour
%-	Note: this is different to rectum where micerectal fat which is more homogeneous as naturally the lung purecuma has more connected tissue septur 

\subsection{Classification Performance}
Results are generated from a mean of 
5-fold cross-validation of MLP classifiers and are summarised in 
Table~\ref{MLP results}. 
Separate results are given for 
1: malignant vs non-malignant and 
2: malignancy vs benign with ambiguous excluded.
This method achieves state-of-the-art results exceeding the 
maximum AUC of 0.967 from \citet{systematic_review} (c.f.~Section \ref{lit_review}). 
For a direct comparison with similar methodology, 
\citet{silva} achieved AUC 0.936 after retraining the encoder. For a comparison to clinical radiologist performance, \citet{AlMohammad2019} conducted a study based on 60 CT scans evaluated by 4 expert radiologists and compared to pathologically confirmed cases. The radiologists had a mean AUC of 0.846, recall of 0.749, specificity of 0.81.
Results provided give performance metrics after initial training and after Expectation-Maximisation optimisation with the classifier loss (`$X_{EM}$'). 
Clearly, the fine-tuning improves the performance of the classifiers 
but also the VAE performance metrics for image reconstruction and the KLD do not significantly change and in most cases improve. The best individual model performance outside of cross-validation is a malignant vs benign classifier using GVAE latent vectors which achieved AUC 0.99 and 95.9\% accuracy. Overall the EM-optimised VAEs had a virtually idenitcal performance, the GVAE had the highest AUC of 0.98 and DirVAE had the highest accuracy of 93.9\%.

% \begin{table}[h!]
% \caption{Latent Vector Malignancy Classifier Results.}
% \vspace{0.3cm}
% \centering
% \begin{tabular}{l l l l l}
% \tophline
% Statistic & G1 (M v B) & G2 (M v ¬M) & D1 (M v B) & D2 (M v ¬M)  \\
% \middlehline
% AUC & 0.89 & 0.85 & 0.84 & 0.83 \\
% Accuracy & 0.82 & 0.79 & 0.77 & 0.78 \\
% Precision & 0.81 & 0.67 & 0.81 & 0.59 \\
% Recall & 0.88 & 0.75 & 0.81 & 0.76 \\
% Specificity & 0.74 & 0.81 & 0.70 & 0.78 \\
% F1 Score & 0.84 & 0.70 & 0.81 & 0.74 \\
% SSIM & 0.63 & 0.66 & 0.59 & 0.53 \\
% MSE & 0.0059 & 0.0082 & 0.0087 & 0.032 \\
% MAE & 0.038 & 0.039 & 0.078 & 0.054 \\
% \bottomhline
% \end{tabular}
% \label{MLP results}
% \end{table} 
%\vspace{-2mm}
\begin{table}[h!]
\caption{Latent vector malignancy classifier results before and after Expectation Maximisation (EM) optimisation. 1: malignant vs non-malignant, 2: malignant vs benign for Gaussian (G) and Dirichlet (D) variants. Results given as $\mu \pm \sigma$ (mean $\pm$ std dev) across 5 runs.}
\vspace{0.2cm}
\centering
\begin{tabular}{l r r r r r r r}
\tophline
Model & AUC & Accuracy & Precision & Recall & Specificity & F1 Score \\
\middlehline
$G1_{EM}$ & \textbf{0.975} $\pm 0.004$ & \textbf{0.934} $\pm 0.014$ & \textbf{0.90} & \textbf{0.92} & \textbf{0.94} & \textbf{0.91} \\
$D1_{EM}$ & 0.974 $\pm 0.001$ & 0.933 $\pm 0.001$ & \textbf{0.90} & \textbf{0.92} & \textbf{0.94} & \textbf{0.91} \\
$G1$ & 0.850 $\pm 0.017$ & 0.793 $\pm 0.018$& 0.69 & 0.74 & 0.82 & 0.71 \\
$D1$ & 0.831 $\pm 0.020$ & 0.782 $\pm 0.030$ & 0.65 & 0.74 & 0.80 & 0.69 \\
\middlehline
$G2_{EM}$ & \textbf{0.980} $\pm 0.008$ & 0.931 $\pm 0.017$ & \textbf{0.93} & \textbf{0.96} & 0.89 & 0.94 \\
$D2_{EM}$ & 0.978 $\pm 0.007$ & \textbf{0.939} $\pm 0.020$ & \textbf{0.93} & \textbf{0.96} & \textbf{0.90} & \textbf{0.95} \\
$G2$ & 0.894 $\pm 0.021$ & 0.819 $\pm 0.027$ & 0.81 & 0.88 & 0.74 & 0.84 \\
$D2$ & 0.841 $\pm 0.013$ & 0.770 $\pm 0.017$ & 0.81 & 0.81 & 0.70 & 0.81 \\
\bottomhline
\end{tabular}
\label{MLP results}
\end{table} 

% give +/- standard devs for each
%One key difference between the two models is that the first model, with ambiguous lesions excluded, was better at correctly detecting malignant tumours with 0.916 precision but in turn was also incorrectly identifying more benign lesions as malignant, with a lower specificity of 0.864. Note that the first model only used 8,494 samples. The maximum AUC for both models was 0.95 and 0.928 respectively and for accuracy 91\% and 89.5\%. The average MSE of correct predictions was 0.0053 and incorrect was 0.0077 suggesting that the  model was better at predicting the malignancy when the VAE reconstructions were better. These results compare directly with \citet{silva} where AUC was 0.936 and the MSE was 0.0016 and 0.0034 respectively. The performance metrics also compare well to other AI based methods built for lung lesion classification \cite{systematic_review}.

%These results compare well to other models built on the LIDC-IDRI dataset, accuracy of 86.8\% in \cite{shen}, accuracy of 91.9\% \cite{svm}, AUC 0.967 and accuracy 89.5\% in \cite{xie}, and finally AUC of 0.936 in \cite{silva} which had a very similar modelling approach.

% add average confusion matrix across cross validation of 10
% bootstrapping could use random.choice? but in general we want 100+ bootstrap samples (too many for this)
%\vspace{-5mm}
\subsection{Clustering and Latent Space Exploration}
%\vspace{-1mm}
%Clustering was performed using both CLASSIX and K-Means for the Gaussian and Dirichlet latent vectors. To decide the cluster size, an Elbow graph of K-Means cluster sizes was evaluated and found only incremental gains after $k=50$, a final number of 88 was chosen. 
\begin{figure}[ht]
\centering
\begin{minipage}[b]{0.4\linewidth}
\centering
\includegraphics[width=\textwidth]{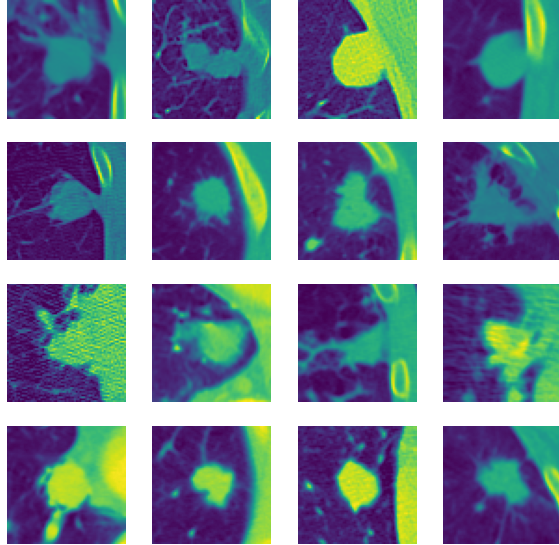}
%\subcaption{(a)}
\end{minipage}
\hspace{0.5cm}
\begin{minipage}[b]{0.4\linewidth}
\centering
\includegraphics[width=\textwidth]{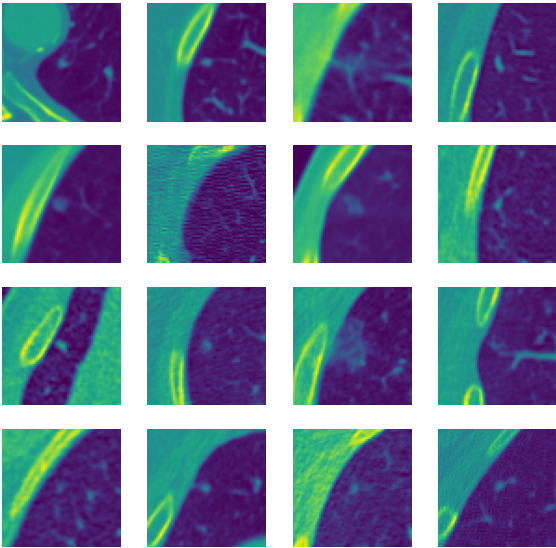}
%\subcaption{(b)}
\end{minipage}
\vspace{0.3cm}
\caption{Latent vectors clustered by visual features: sample of 16 images from a cluster with 100\% malignant lesions (a) and a cluster with 97\% non-malignant lesions (b).}
\label{clusters}
\end{figure}
In Figure \ref{clusters} visual similarities can be observed, for instance in (a) there is a large circular mass in the centre, whereas in (b) more bone is concentrated in the top left corner.

\begin{figure}[ht]
\centering
\begin{minipage}[b]{0.4\linewidth}
\centering
\includegraphics[width=\textwidth]{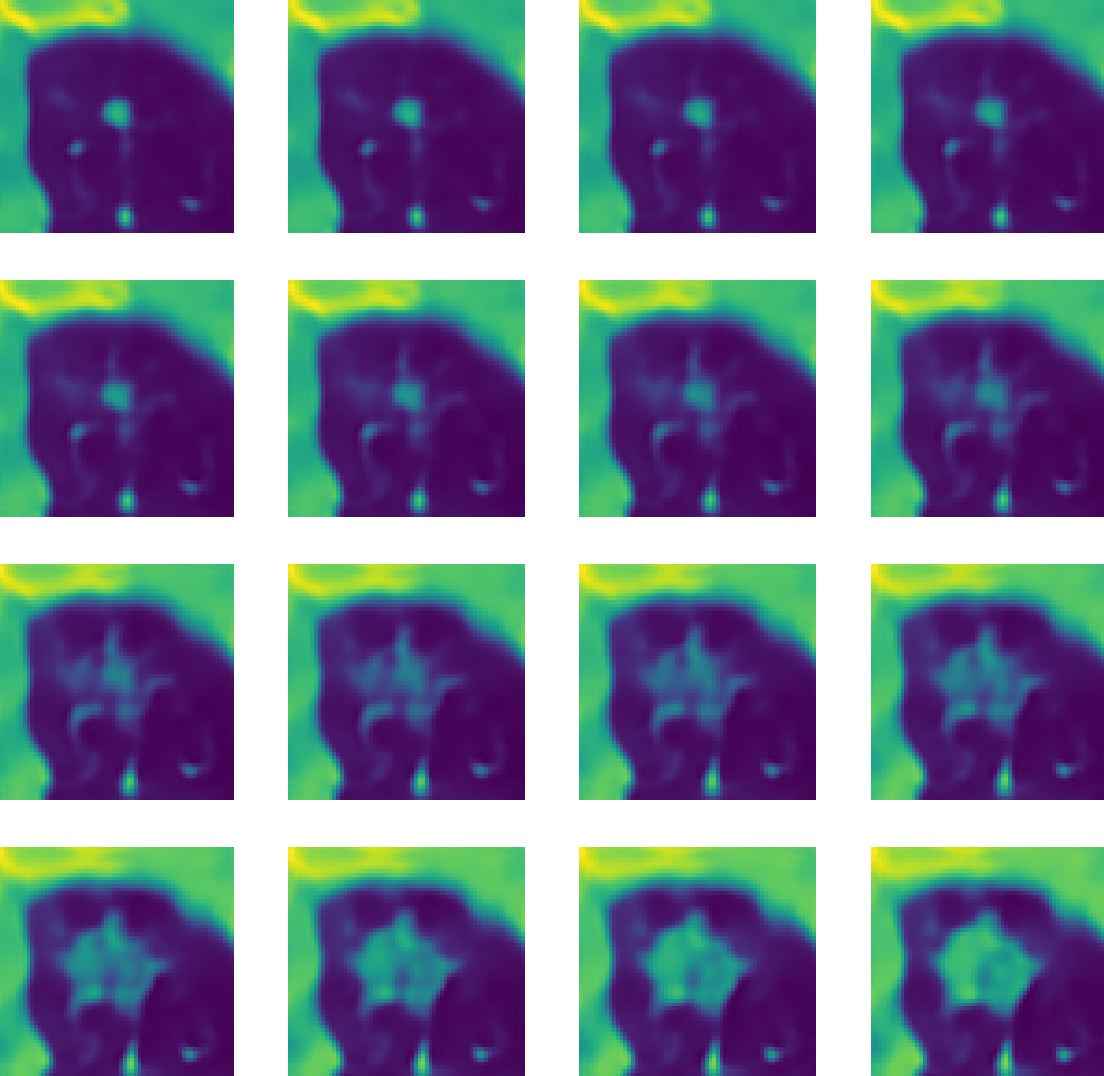}
%\subcaption{(a)}
\end{minipage}
\hspace{0.5cm}
\begin{minipage}[b]{0.4\linewidth}
\centering
\includegraphics[width=\textwidth]{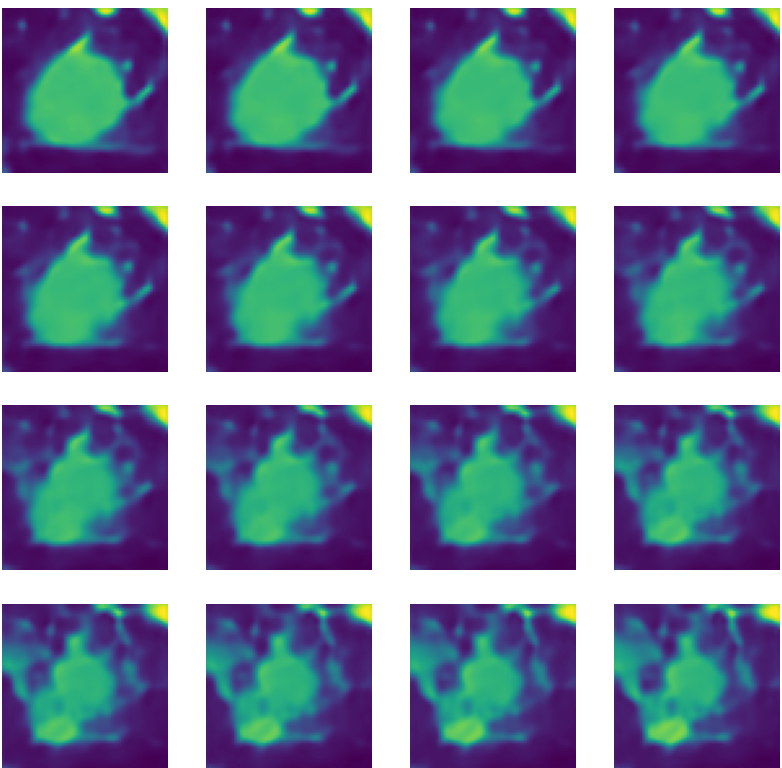}
%\subcaption{(b)}
\end{minipage}
\vspace{0.3cm}
\caption{Capability for clinically meaningful traversals in the latent space related to lung tumour growth (a) and increased parenchyma/irregular tumour border (b) (c.f.~Section \ref{section_clustering_explanation}).}
\label{tumour_growth_law}
%\vspace{-5mm}
\end{figure}

% \begin{table}[h!]
% \caption{CLASSIX and K-Means Clustering Statistics}
% \vspace{0.2cm}
% \centering
% \begin{tabular}{l l l l l}
% \tophline
% Statistic & $G_C$ & $D_C$ & $G_{KM}$ & $D_{KM}$  \\
% \middlehline
% Patients in a one cluster & 20\% & 25\% & 20\% & 19\% \\
% 50\% slices in one cluster & 48\% & 58\% & 43\% & 44\% \\
% 25\% slices in one cluster & 90\% & 94\% & 87\% & 87\% \\
% Clusters with above 75\% of one class & 48\% & 58\% & 44\% & 44\% \\
% Clusters with above 66.67\% of one class & 73\% & 72\% & 65\% & 63\% \\
% \bottomhline
% \end{tabular}
% \label{Clustering results}
% \end{table} 
\begin{table}[h!]
\caption{CLASSIX (C) and K-Means (KM) clustering statistics. Patient abbreviated as Pt.}
\vspace{0.2cm}
\centering
\begin{tabular}{l l l l l}
\tophline
Statistic & $G_C$ & $D_C$ & $G_{KM}$ & $D_{KM}$  \\
\middlehline
Pt in a single cluster & 13\% & \textbf{25\%} & 15\% & 19\% \\
50\% Pt slices in one cluster & 33\% & \textbf{51\%} & 35\% & 42\% \\
25\% Pt slices in one cluster & 81\% & \textbf{91\%} & 80\% & 84\% \\
Clusters with above 75\% of one class & 66\% & \textbf{77\%} & 75\% & 63\% \\
Clusters with above 66.67\% of one class & 82\% & \textbf{88\%} & 86\% & 78\% \\
\bottomhline
\end{tabular}
\label{Clustering results}
%\vspace{-2mm}
\end{table} 
Clustering statistics for the GVAE (G) and DirVAE (D) models with 131 clusters are given in Table~\ref{Clustering results};
these show that the latent space is capable of separating the lesions based on
clinically relevant features such as tumour size and malignancy class, and furthermore
attempts to group multiple images of the same patient together. 
%The DirVAE latent vectors have a better separation according to these statistics which may infer the latent space has a more disentangled representation. It is worth noting that when comparing the above statistics before and after VAE retraining with classifier loss, the separation based on malignancy class increases and by patient decreases. This indicates that the classifier loss encourages the VAE to encode features corresponding to class into the latent space and hence lesions of the same class are placed closer together in the space. 
It is worth noting that this clustering is post EM optimisation, which increased the separation by malignancy class. This indicates that the VAE was encouraged to encode features related to class in the latent space. Although, the clusters already had a high separation before using the classifier loss which indicates the latent space naturally encodes these meaningful attributes.
\\ \\ \\ \\
Finally, to demonstrate the capabilities of VAE models in this domain, in Figure \ref{tumour_growth_law} there are two
examples of latent space traversals (c.f.~Section \ref{section_clustering_explanation}). These directions were applied to a new lesion not used in finding the direction and it appears to generalise well including maintaining the surrounding bone structure and generating realistic images at each step. Animations of latent traversals were generated by this analysis showing smooth transitions with more samples. Traversals are constructed by sampling from the latent space, either by using a start and end image and interpolating, or choosing a start point and moving in the direction of the desired feature as in Figure \ref{tumour_growth_law}. Note that all images other than the start point are synthetic. Further examples are provided on the GitHub page.

% \begin{figure} %[h!] 
% \centering % change to 4.5
% \includegraphics[width=0.4\textwidth]{Images/tumour_bigger_gif.png}
% \vspace{0.3cm}
% \caption{Application of a direction vector in latent space related to tumour growth applied to a new image.}
% \label{tumour_growth_law}
% \vspace{-5mm}
% \end{figure}

% ideas:
%Proportion of tumour vs other lesion in each cluster bar chart over random sample of 10 clusters or all of them
%e.g. is hierarchical 1st split cancer/clean (non-zero chance?)

%\vspace{-4mm}
\section{Discussion}
%\vspace{-2mm}
The most significant contribution of this work is the novel use of DirVAEs in the cancer imaging domain. This work has also shown that a VAE and MLP combination can achieve state-of-the-art classification performance for lung lesion diagnosis with AUC 0.98 which compares to radiologist performance of 0.846 and is on par with the best AI-based approaches. Overall the results suggest that both approaches produce good classification models, the key difference is that the DirVAE demonstrates greater disentanglement and separation by clinically meaningful characteristics, whilst GVAE produces better reconstructions. In practice, the best model will likely depend upon the context, dataset and specific task.

This approach for encoding the images with a VAE lends robustness and an element of explainability as we can observe that lesions with similar characteristics have representations that are close together in the latent space as demonstrated by the clustering results. This aspect of the work may be valuable for generating pseudo-labels in tasks without a ground truth.
%the DirVAE has better disentanglement
Although this paper demonstrates accurate classification models, it is important to discuss some of the limitations of the proposed method. 
Firstly, the labels are generated by expert radiologists rather than the gold standard of pathological confirmation. 
Secondly, the data uses a non-standardised slice thickness, while some may argue it is better to standardise, this approach may be more generalisable to the real world. 
%where models could encounter non-standardised slices from different scanners. 
One further limitation of the 2D approach is that slices from the same patient are not independent both in structure and the likelihood of malignancy. While extending this analysis to 3D may produce a more robust model, data samples would reduce from 13,852 to 875 and model complexity would increase. 

% What about relying on pre-specified bounding boxes, so dependent upon a clinician / other model to do this first.

Some of the lung parenchyma were not fully captured by the latent vectors as demonstrated in Figure~\ref{vaerecon}.
However,
the lung naturally has more connective tissue septa than other 
parts of the body and these hold little relevance to malignancy diagnosis,
meaning that failure to capture the parenchyma could actually 
increase the signal-to-noise ratio.
Further experimentation is needed to determine whether they are important for the overall classification.
%However, it is difficult to tell whether the details the VAE is not capturing are normal parenchyma or information crucial for a malignant lesion diagnosis.

%    \item comment on novelty - explored latent space but more can be done with finding direction which makes tumour bigger or adds trabeculations which may increase or reduce probability of malignancy 
%    \item resampling CT scans, pixel spacing was set to [1.00, 1.00, 1.00] mm and each CT dimension was calculated to match this new spacing, obtaining the resampled by interpolation \cite{silva}
%    \item classifying ambiguous as benign is not good!
%\end{itemize}

%\vspace{-2mm}
\section{Conclusion and Future Work}
%\vspace{-1mm}
Overall, (1) VAEs with Gaussian and Dirichlet priors were trained to produce a latent space which was capable of capturing macro details to a very high standard and micro details to a satisfactory standard. (2) Clustering algorithms were implemented, with results showing that latent vectors were clustered by patient and lesion type and that the Dirichlet prior was better at separating the data in this way. (3) MLP classifiers for malignant or benign lesions were trained using latent vectors from the VAEs, the best model achieved state-of-the-art performance with an AUC of 0.98 and 93.1\% accuracy.

Future work could include combining 2D slice level prediction into higher level predictions such as at the 3D lesion or patient level. This would mitigate the limitations associated with a 2D approach including slice thickness and independence of samples. Further improvements to the VAE methodology could include segmenting bone and fat to remove this impact from the latent space. Additionally, extending the latent space exploration to see how different features affect classifications. For instance, using the tumour growth direction or other feature changes such as adding/removing parenchyma to see the impact on the probability of malignancy. Applying methods for latent direction discovery by selecting the best traversals based on metrics such as largest change in prediction score. Finally, to look at implementing DirVAE latent traversals along single dimensions to demonstrate its disentanglement and to add value for model interpretation.

%\begin{itemize}
%    \item The British Machine Vision Conference (BMVC): Paper deadline Friday, 12 May 2023 (20th - 24th November 2023, Aberdeen) (Maximum 9 pages) \url{https://bmvc2023.org/}
    %\item Paper on VAEs from last year BMVC \url{https://bmvc2022.mpi-inf.mpg.de/0636.pdf} 
    %\item not sure how to limit the number of authors so can just delete them and write et al. - is 5 authors a good amount?
%\end{itemize}

% \begin{table}[h!]
% \vspace{0.3cm}
% \centering
% \begin{tabular}{l r r r r r r}
% \tophline
% & VAE$_{EM}$ & VAE & Systematic Review [3] & Silva et al. [4] & Radiologist [1] \\
% \middlehline
% AUC & \textbf{0.98} & 0.89 & 0.7 - 0.97 & 0.94 & 0.85 \\
% Accuracy & \textbf{0.93} & 0.82 & 0.88 - 0.99 & 0.90 & \\
% \bottomhline
% \end{tabular}
% \label{MLP results}
% \end{table} 

\import{./}{bibliography.tex}
\newpage
\end{document}

%% file: bibliography.tex
\bibliography{bibliography}
\bibliographystyle{unsrt}

\citestyle{acmauthoryear}